\documentclass[11pt,a4paper]{article}
\usepackage[hyperref]{acl2020}
\usepackage{times}
\usepackage{graphicx}
\usepackage{latexsym}

\usepackage{microtype}

\aclfinalcopy % Uncomment this line for the final submission
\title{TicketTalk: Toward human-level performance with end-to-end, transaction-based dialog systems}

\author{Bill Byrne*, Karthik Krishnamoorthi*, Saravanan Ganesh*, Mihir Sanjay Kale\\
\textit{Google, Mountain View, CA} \\
\texttt{\{billb,krishnamoorthi,srrvnn,mihirkale\}@google.com}
}

\date{}

\begin{document}
\maketitle
\def\thefootnote{*}\footnotetext{Equal contribution}\def\thefootnote{\arabic{footnote}}
\begin{abstract}
We present a data-driven, end-to-end approach to transaction-based dialog systems that performs at near-human levels in terms of verbal response quality and factual grounding accuracy.
We show that two essential components of the system produce these results: a sufficiently large and diverse, in-domain labeled dataset, and a neural network-based, pre-trained model that generates both verbal responses and API call predictions.
In terms of data, we introduce TicketTalk, a movie ticketing dialog dataset with 23,789 annotated conversations.
The movie ticketing conversations range from completely open-ended and unrestricted to more structured, both in terms of their knowledge base, discourse features, and number of turns.
In qualitative human evaluations, model-generated responses trained on just 10,000 TicketTalk dialogs were rated to “make sense” 86.5\% of the time, almost the same as human responses in the same contexts.
Our simple, API-focused annotation schema results in a much easier labeling task making it faster and more cost effective.
It is also the key component for being able to predict API calls accurately. We handle factual grounding by incorporating API calls in the training data, allowing our model to learn which actions to take and when.
Trained on the same 10,000-dialog set, the model’s API call predictions were rated to be correct 93.9\% of the time in our evaluations, surpassing the ratings for the corresponding human labels.
We show how API prediction and response generation scores improve as the dataset size incrementally increases from 5000 to 21,000 dialogs.
Our analysis also clearly illustrates the benefits of pre-training.
To facilitate future work on transaction-based dialogs, we have publicly released the TicketTalk dataset at \url{https://git.io/JL8an}.

\end{abstract}

\section{Introduction}

Building a dialog system that handles human conversational behavior is challenging because it must respond sensibly and relevantly to a wide variety of context-sensitive user input over multiple conversation turns. Task-based systems, e.g. those used for ticket booking, food ordering, etc., face further hurdles to incorporate ever changing, real-world knowledge into the dialog and execute transactions. Recently, there has been growing interest in the so-called end-to-end approach to task-based dialog systems \cite{peng2020soloist,hosseini2020simple,lin2020mintl,wen-etal-2017-network,bordes2016learning} due to its relatively simple and scalable architecture, and promising results in chatbot applications \cite{vinyals2015neural,serban2015hierarchical}. Inspired by sequence-to-sequence learning \cite{sutskever2014sequence}, this approach trains a single model on a dialog dataset to form the basis for a given application. For each dialog turn, the model effectively takes the conversation history as its input and generates an appropriate response.

To gain wider adoption, the end-to-end approach must overcome challenges with respect to training data and factual grounding. In terms of training data, there is already general concern in the NLP community about the lack of quality, task-oriented dialog datasets, especially domain-specific collections \cite{wen-etal-2017-network,bordes2016learning}. This problem is compounded for end-to-end approaches since they typically require a large amount of in-domain data to generate competitive results. With respect to grounding, since the end-to-end approach is based on a single neural network, it must either incorporate the knowledge base (KB) into the model itself, or the model must be able to accurately predict which API calls to make and when. In addition, details returned from the API calls must be accurately incorporated in conversational responses. This is contrasted with modular architectures where the user’s intent is derived from a structured representation and then used to determine which API calls to make such as in \citet{rastogi2020towards} and \citet{madotto2020language}.

In this work we promote an end-to-end approach to single-domain, transaction-based dialog systems and describe how we overcome both data and grounding challenges described above. In qualitative evaluations, our models perform on par with humans in generating verbal responses as well as predicting API calls. Just two components form the basis for this system: a sufficiently large, in-domain, labeled dataset and a pre-trained transformer model. Combining natural language output and structured API calls into a unified text-to-text-format allows us to leverage general purpose text-to-text transformers to train models. Specifically, we use the T5 infrastructure \cite{raffel2019exploring} and show that its pre-training feature has a significant impact on evaluations, boosting scores by 30 percent.

Models were trained on our TicketTalk dataset, our new movie ticketing dialog dataset with 23,789 conversations labeled with a simple yet unique API-based annotation schema. This makes it one of the largest, single-domain datasets to date. A public release of the dataset accompanies this paper. We chose movie ticketing since it is both transaction-based and relatively complex, but our overall approach to dialog systems applies to any task-based domain. While there is a lot of recent work on multi-domain task-based dialog systems, human-like interaction for even single-domain tasks has yet to be demonstrated. By first solving the problem for a single domain, we argue that replicating the process for multiple domains will be achievable by simply training additional high-quality datasets labeled with the same API-focused strategy.

\section{Related work and background}
\subsection{Datasets}

Over the past few years the NLP community has responded to the lack of dialog data with larger, publicly released task-oriented datasets spanning multiple domains \cite{wu2020tod,budzianowski2019hello}. This underscores the crucial role data plays in any approach to task-based dialog systems. MultiWOZ \cite{budzianowski2018multiwoz} consists of 10,420 dialogs in multiple domains and has become a popular benchmarking corpus for state tracking. It has also undergone a series of subsequent refinements. MSR-E2E, featured in the Microsoft dialog challenge \cite{li2018microsoft}, has 10,087 dialogues in three domains, movie-ticket booking, restaurant reservation, and taxi booking. \cite{byrne2019taskmaster} offers 13,215 dialogs in six domains and has been updated with a second installment, Taskmaster-2 \cite{Taskmaster2}, which adds 17,289 more dialogs totalling over 30,000. The Schema Guided Dialogue dataset \cite{rastogi2020towards} has 22,825 dialogs in multiple domains. MetaLWOZ \cite{lee2019multi} has 37,884 dialogs in 227 domains and is aimed at helping models more accurately predict user responses in new domains. Both Schema and MetaLWOZ are used in DSTC8 \cite{DSTC8}. In addition to these, \citet{serban2018survey} provides a thorough survey of dialog corpora released in previous years.

\subsection{Modular vs. end-to-end architectures}

In contrast to the end-to-end \footnote{The term “end-to-end” is sometimes also used when describing parts of modular systems \cite{li2017end,wen-etal-2017-network} but it is fundamentally different from the single text-to-text transformer model approach we present here.} approach, traditional, modular strategies employ a division of labor among the components, e.g. understanding, state tracking, dialog policy, generation, etc., which are either largely hand-crafted or derived from training individual models on labeled datasets \cite{wen-etal-2017-network,young2013pomdp}. This architecture is inherently more complex than the single-model end-to-end strategy we propose and can require significantly more design and engineering. Moreover, since each module requires its own supervised training dataset, it is harder to apply to different domains \cite{serban2015building}.

\begin{figure}[h]
\centering
\includegraphics[width=0.48\textwidth]{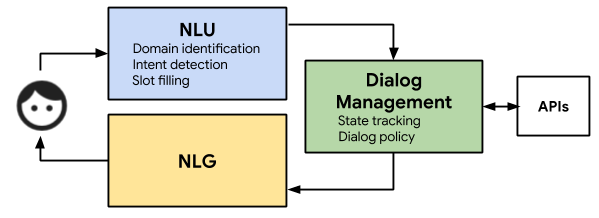}
\caption{Traditional modular system}
\label{fig:modular}
\end{figure}

However, the separation of functions makes the modular approach more transparent and in some respects easier to debug. It has also been considered by some to be better equipped to interact with external APIs \cite{sukhbaatar2015end,wen-etal-2017-network} and therefore might be better suited for task-based dialogs. As mentioned above, we show that our single model-based approach can accurately generate both the appropriate response as well as predict the correct API call at the right time.

\begin{figure}[h]
\centering
\includegraphics[width=0.40\textwidth]{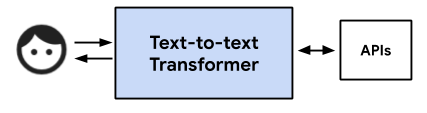}
\caption{Simplified end-to-end system}
\label{fig:E2E}
\end{figure}

\section{The TicketTalk dataset}
\subsection{Overview}

The TicketTalk movie ticketing dataset was created using the self-dialog collection method \cite{krause2017edina, moghe2018towards,byrne2019taskmaster} where a crowd-sourced worker writes both sides of the dialog (i.e. both customer and ticketing agent turns) based on a particular scenario and set of instructions. Following the annotation strategy used for Taskmaster-1 \cite{byrne2019taskmaster}), we limit labels to basic entities and events (i.e. API calls).

\begin{table}[h]
\renewcommand{\arraystretch}{1.3}
\centering
\setlength{\tabcolsep}{13pt}
\begin{tabular}{lrl}
\hline \textbf{\textsf{STAT TYPE}} & \textbf{\textsf{VALUE}} \\ \hline
Dialogs & 23,789\\
Total turns & 481,632\\
Unique tokens & 62,868 \\
Avg. turns per dialog & 20.25\\
Avg. tokens per turn &  10.35\\
Unique named entities & 57,285\\
\hline
\end{tabular}
\caption{\label{font-table} TicketTalk Dataset Statistics }
\end{table}

The rationale for limiting dialogs to a single domain (movie ticketing) is based on our hypothesis that human-level performance in terms of both response generation and API call prediction for a particular task requires larger (i.e. 10,000+), more diverse datasets than are currently available. In other words, carefully curated, annotated datasets that cover all the idiosyncrasies of a single task or transaction are a key factor in model performance. Concern about the cost and efficiency of creating these larger corpora has led some researchers to look for approaches that alleviate dependencies on annotated data \cite{budzianowski2019hello,wen-etal-2017-network}. However, significant time and expense can be saved when assembling these corpora by simplifying the collection and annotation procedures. In addition, little to no training is required for workers to be able to perform consistently well.

\subsection{Collection methodology}

Using self-dialogs (where a worker creates the whole conversation, both user and agent turns) facilitates building large and linguistically rich datasets since it is both simple and cost effective, and allows users to draw on their lifetime of conversational experiences. This in turn ensures the model can handle the wide range of human conversational behaviors that emerge in natural dialog. For this project we extended the self-dialog to include over three dozen sets of user instructions to generate a wider variety of conversations, from open-ended prompts to more specific instructions that require specific types of exchanges. For example, one set simply instructs workers to “write the transcription of a conversation” in which a person makes a successful ticket transaction with a booking agent. This allows dialog creators to express their unique view of what a typical movie ticketing transaction would be, structuring each conversation how they see fit. They are also instructed to find real values for required details (i.e. slots) such as  time, date, theater, movie, etc. using a movie or theater site of their choice for a specific location. This ensures the dataset has a large and diverse KB. In contrast, the more restrictive sets of instructions focus on specific sub-dialogs for error handling, changing a detail, entity resolution, and the like. In such cases we often provide a limited KB with one or more values for all the details so the worker can focus on the primary task of creating a realistic set of exchanges for this type of interaction. In a third type of scenario, the conversation is partially completed and the user’s task is focused on a very specific part of the exchange. This allows us to “fill holes” in the data quickly and cost effectively. That is, we can create large numbers of short, conversational examples that the model does not handle adequately and then retrain for better results.

\subsection{Annotation}

Dialog data annotation can be complex and time consuming even for trained linguists as it typically involves carefully and consistently labeling dialog states, user intents, and dialog acts, among other possible labels \cite{henderson2013deep,wen-etal-2017-network,budzianowski2018multiwoz}. The API-targeted approach is far more straightforward since only basic entities (e.g. name, time, number of tickets, theater, movie attributes, etc.) and API calls (e.g. to find theaters, movies, and showtimes, book tickets, etc.) are labeled. The task is therefore easier to learn, faster to complete, and cheaper to run. Moreover, as we discuss below, it fits well with the text-to-text format we use in our approach to transaction-based dialog systems. The full annotation schema is included with the dataset release.

\section{A novel end-to-end approach}
\subsection{Overview}

We implement a new approach to end-to-end dialog systems by combining natural language output and structured API calls into a unified text-to-text format where the input and output are always text strings. This allows us to leverage widely available, state of the art, general purpose text-to-text transformers as the foundation of our system. Specifically, we used the publicly available Text-To-Text Transfer Transformer (T5) \cite{raffel2019exploring} to train our models. The T5 framework was designed specifically to explore transfer learning techniques for NLP and includes pre-training on the Colossal Clean Crawled Corpus (C4), composed of hundreds of gigabytes of web-based English text \cite{raffel2019exploring}. The original pre-training objective for the C4 corpus in the T5 framework was a denoising task, i.e. recovering missing words from the input. Since this type of task scales well to multiple downstream tasks, we used our custom inputs/targets from the TicketTalk dataset to represent an end-to-end task based dialog system and ultimately achieve  positive results.

\subsection{Setup}
We use T5-Base \cite{raffel2019exploring} as our pre-trained model, which follows the transformer architecture \cite{vaswani2017attention}  and consists of 220M parameters. It was pre-trained on the large scale C4 dataset mentioned above for 1M steps with a span corruption objective. We fine-tune this model on the Taskmaster-3 dataset for 40000 steps with a constant learning rate of 0.001 using 16 TPU v3 chips. The batch size was set to 131,072 tokens per batch. The maximum input sequence length and output length were set to 1024 and 256 tokens respectively.

\subsection{Model and implementation}

The goal of our model is to generate a text string that either serves as a verbal response to the user or that contains one or more API calls with the data required at the current stage of the conversation. Verbal responses come in two flavors: those that depend on a particular API call details and those that do not. For example, when an API is invoked to find theater names for a given movie and location, the details returned from the API call must be correctly incorporated into the system’s next response, e.g. “I found two theaters, AMC 20 and Century City 16.” In contrast, other verbal outputs, e.g. “What city do you plan to see the movie in?” are derived from the overall conversation history.

Given the required text-to-text format used in our approach, we identify the type and function of each string by converting the annotations to a set of tokens. As shown in Table \ref{table:tokens}, tokens identify the speaker, i.e. user vs. agent, the string type i.e. utterance vs. API call, and the details of each API call, both names as well as input parameters and values, and response parameters and values. We also tag the conversation “context” which separates the most recent turn from previous turns. Our token key is shown in Table \ref{table:tokens}.

\begin{table}
\small
\centering
\renewcommand{\arraystretch}{1.3}

\setlength{\tabcolsep}{6pt}
\begin{tabular}{ll}
\hline 
\textbf{U} & user\\
\textbf{A} & agent\\
\textbf{PN} & program name \\
\textbf{PAN} & program argument name\\
\textbf{PAV} &  program argument value\\
\textbf{PR} & program response\\
\textbf{PRAN} & program response argument name\\
\textbf{PRAV} & program response argument value\\
\textbf{C} & conversation context\\
\hline
\end{tabular}
\caption{\label{table:tokens} \small{Tokens identifying string type and function}}
\end{table}

The first step is to use tokens to represent the user and agent interactions, providing speaker information to the model by the use of "$<$U$>$" and "$<$A$>$". We then convert any API invocations into their text equivalent using tokens for marking API names, argument types and values, i.e. "$<$PN$>$", "$<$PAN$>$", etc. The results of these two steps are shown in Table \ref{table:speaker-tokens}.

\begin{table}[!ht]
\small
\sffamily
\renewcommand{\arraystretch}{1.3}
\setlength{\tabcolsep}{6pt}
\begin{tabular}{ll}
 \textbf{$<$U$>$}&I’d like to watch a movie.\\
    \textbf{$<$A$>$}&Sure. I can help you with that.\\
    & What kind of movies are you interested in?\\
    \textbf{$<$U$>$}&Are there any good action movies?\\
&\textbf{$<$PN$>$}find\_movies\\
&\textbf{$<$PAN$>$}name.genre\textbf{$<$PAV$>$}action\\
&\textbf{$<$PR$>$}find\_movies\\
&\textbf{$<$PRAN$>$}name.movie\\
&\textbf{$<$PRAV$>$}John Wick\textbf{$<$PRAV$>$}Jack Ryan\\
    \textbf{$<$A$>$}&I found John Wick and Jack Ryan.\\
\end{tabular}
\caption{Speaker turns and API calls identified with tokens}
\label{table:speaker-tokens}
\end{table}

The next step is to create the model inputs and targets. We use the following algorithm to accomplish this:
\begin{enumerate}
    \item Initialize conversation context to an empty string.
    \item Iterate through the interactions and do the following:
    \begin{enumerate}
        \item If the sentence is a user utterance ($<$U$>$) or a program response($<$PR$>$), add it to the model input along with the conversation context (if present). 
        \item If the sentence is an agent utterance ($<$A$>$) or program invocation ($<$PN$>$), add it to the model target.
        \item If both model input and target have been created, output the (input, target) pair and update the conversation context to reflect this.
        \item Continue (2) to generate the next input, target pair.
    \end{enumerate}
\end{enumerate}

Using the these rules, the model inputs and targets are generated as in Table \ref{table:input-target}.

\begin{table}[h]
\footnotesize
\sffamily
\renewcommand{\arraystretch}{1.4}

\begin{tabular}{|p{33mm}|p{33mm}|}
\hline \textbf{\textsf{INPUTS}} & \textbf{\textsf{TARGETS}}\\ \hline 
\textbf{$<$U$>$}I’d like to watch a movie. &\textbf{$<$A$>$}Sure. I can help you with that. What kind of movies are you interested in? \\\hline
\textbf{$<$U$>$}Are there any good action movies?

\textbf{$<$C$>$}

\textbf{$<$U$>$}I’d like to watch a movie.

\textbf{$<$A$>$}Sure. I can help you with that. What kind of movies are you interested in? &\textbf{$<$PN$>$}find\_movies \textbf{$<$PAN$>$}name.genre \textbf{$<$PAV$>$}action\\
\hline
\textbf{$<$PR$>$}find\_movies \textbf{$<$PRAN$>$}name.movie \textbf{$<$PRAV$>$}John Wick \textbf{$<$PRAV$>$}Jack Ryan

\textbf{$<$C$>$}

\textbf{$<$U$>$}I’d like to watch a movie.

\textbf{$<$A$>$}Sure. I can help you with that. What kind of movies are you interested in? \textbf{$<$U$>$}Are there any good action movies? \textbf{$<$PN$>$}find\_movies \textbf{$<$PAN$>$}name.genre \textbf{$<$PAV$>$}action & \textbf{$<$A$>$}I found John Wick and Jack Ryan.\\

\hline

\end{tabular}
  \caption{\small{Generating inputs vs. targets}}
  \label{table:input-target}
\end{table}

Once the model has been trained on inputs and targets, we can use the system to accomplish tasks in the following manner:
\begin{enumerate}
    \item Obtain user utterance and format it by adding the speaker token.
    \item Provide the formatted utterance to the model.
    \item Obtain model prediction
    \begin{enumerate}
        \item If the model prediction contains the agent ($<$A$>$) token, format it and show it to the user.
        \begin{enumerate}
            \item Update conversation context and start again from (1).
        \end{enumerate}
        \item If the model prediction contains the program ($<$PN$>$) token:
        \begin{enumerate}
            \item Extract program argument name ($<$PAN$>$) and value ($<$PAV$>$).
            \item Issue the API call by providing it to the API adapter.
            \item Format API results and provide it to the model along with the conversation context.
            \item Start from (3).
        \end{enumerate}
    \end{enumerate}
\end{enumerate}

\begin{figure}[h]
\centering
\includegraphics[width=0.35\textwidth]{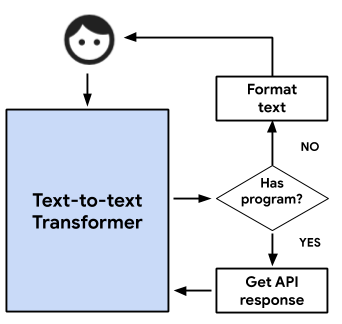}
\caption{System interaction life cycle}
\label{fig:lifecycle}
\end{figure}

This interaction lifecycle is illustrated in Figure \ref{fig:lifecycle}.

\subsection{Invoking APIs}

\begin{figure}[h]
\centering
\includegraphics[width=0.45\textwidth]{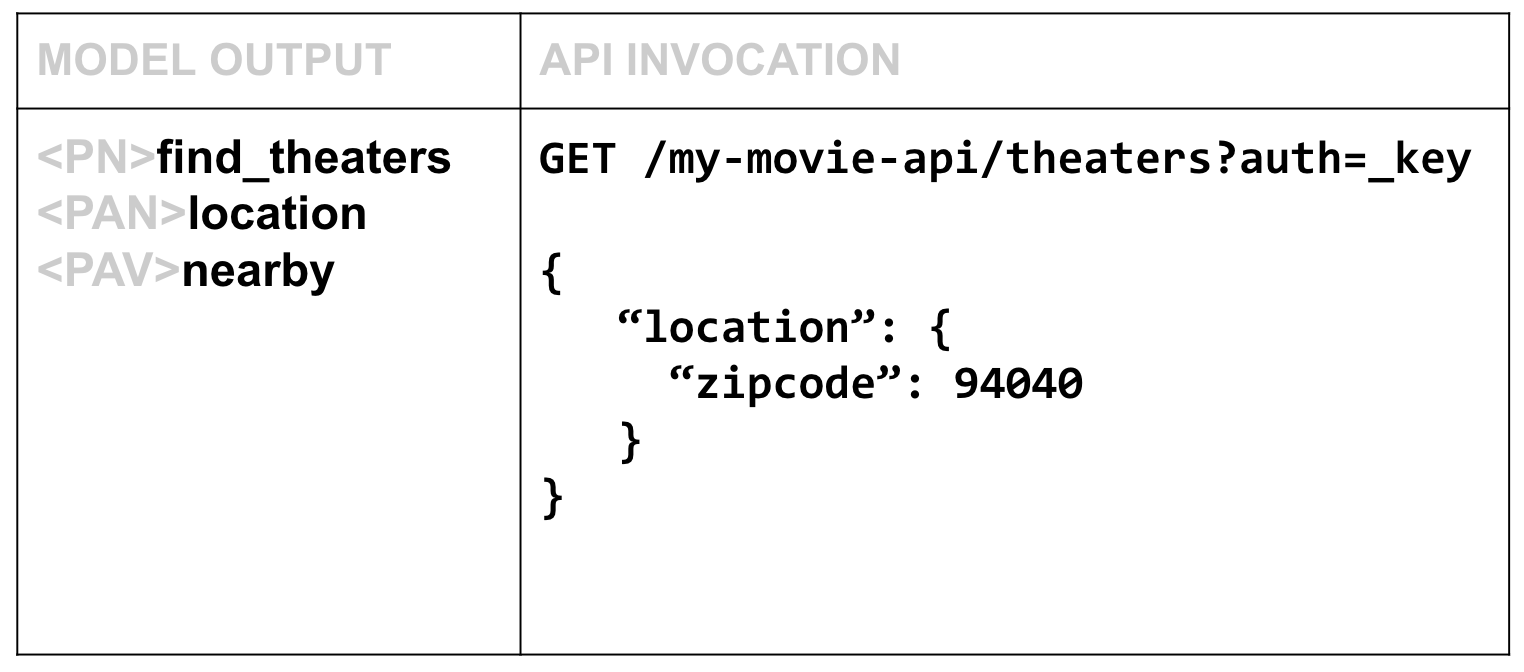}
\caption{Example API invocation (outside model)}
\label{fig:invoke-api}
\end{figure}

When we detect an API call in the output, we invoke the API, retrieve the results, and embed the responses in the next model input. As shown in Figure \ref{fig:invoke-api}, each API call predicted by the model typically contains a generic API name, such as "find-movies", or "find-theaters", and a list of key value pairs that detail the specific parameters to be used while invoking the API, as shown in Figure \ref{fig:invoke-api}.

The API call, while structured, may still include pronouns or other co-referential phrases as input parameters. For example, the date parameter for an API call might contain the value “tonight”, and the location value might be “nearby”. The resolution of these entities happens outside the core interaction layer in what can be understood as the “API adapter” (and not the actual API itself). This not only helps simplify annotation, but also helps leverage existing solutions to these well defined problems. This separation of the API layer is also useful for encapsulating all API specific artifacts, like authentication tokens, endpoint addresses and data formatters. In this way, the end-to-end system is able to interact with the user to solicit details relevant to the task, generate API calls to fetch data from external knowledge sources, and use the responses provided by the API call to construct natural language responses.

\section{Experiments}
\subsection{Overview}

In this section, we show how our end-to-end approach to transaction-based dialog systems produces verbal responses and predicts API calls with near human-level quality and accuracy. Through human qualitative evaluations, we show that two aspects in particular, dataset size and pre-training, significantly affect performance. Below we describe our evaluation methodology followed by a detailed discussion of the experiment results.

\subsection{Evaluation methodology}

Dataset size and pre-training are key factors in creating models for end-to-end dialog systems. To understand the amount of data required for our approach, we trained four models, each on a different number of randomly selected subsets of the TicketTalk dataset, namely 5000, 7500, 10,000 and 21,000 dialogs. To measure the effect of transfer learning, we trained a second 10,000-dialog model without the T5 framework’s pre-training component, setting up an A-B comparison with the pre-trained model.

As mentioned earlier, our models generate three types of output: API calls, verbal responses based on the results of an API call, and “plain” verbal responses based on the conversation context (i.e. not dependent on a particular API call response). We set up a pair of evaluations for each type. The first evaluation asked human raters to evaluate the model’s output given a specific conversation history (i.e. context) while the second asked raters to evaluate the human’s response for the same set of contexts. Each experiment included 1000 context-response pairs of varying lengths, i.e. some conversation histories might have just one exchange (a user and agent turn) while others could have up to nine exchanges. We requested three ratings for each question distributed among a pool of about 900 paid raters for a total of 3000 data points per experiment. Table \ref{table:ewok-plain-pair} and Table \ref{table:ewok-api-pair} below shows a sample context-response pair presented to human raters for each type of model output.
\begin{table}[h]
\sffamily
\footnotesize
\renewcommand{\arraystretch}{1.3}
  \centering
  \begin{tabular}{ |p{33mm}|p{33mm}| }
\hline \textbf{\textsf{CONTEXT}} & \textbf{\textsf{NEXT RESPONSE}}\\ \hline 
\textbf{Cust:} Can you help me book a movie ticket? & \textbf{Agent:} OK. Do you have any theaters in mind?  \\
\textbf{Agent:} Yes I can. & \\
\textbf{Cust:} Can you find tickets for the movie Knives Out? & \\
\textbf{Agent:} Sure! What time did you want to book? & \\
\textbf{Cust:} 5 PM would be best. & \\
\hline
  \end{tabular}
  \caption{\small{Context paired with generated verbal response}}
  \label{table:ewok-plain-pair}
\end{table}

\begin{table}[h]
\sffamily
\footnotesize
\renewcommand{\arraystretch}{1.3}
  \centering
  \begin{tabular}{ |p{35mm}|p{33mm}|  }
\hline \textbf{\textsf{CONTEXT}} & \textbf{\textsf{ACTION}}\\ \hline 
\textbf{Cust:} I would like to see a movie tonight.  &  FIND\_MOVIES location: Oak Valley Arkansas \\
\textbf{Agent:} Sure. What movie would you like to see? & \\
\textbf{Cust:} I'm not really sure. Can you help me pick something?&\\
\textbf{Agent:} No problem. I can give you the names of a couple of movies playing in your area. What city are you going to see the movie in?&\\
\hline
  \end{tabular}
  \caption{\small{Context paired with predicted API call}}
  \label{table:ewok-api-pair}
\end{table}

We use our “makes-sense” metric to evaluate the model-generated responses and API call predictions against the human standard. For verbal responses, we ask one question:
\begin{itemize}
    \item Does the agent's next response make sense?
\end{itemize}
For negative answers, we give a list of reasons raters believe it does not make sense (i.e. off topic, repeated information, incorrect details, language mistakes, other).
For API call predictions there are two questions:
\begin{enumerate}
    \item Do all the action types, their details, and their order make sense at this point in the conversation?
    \item Are there any actions that should be listed here but that are missing (either as additions or replacements)? 
\end{enumerate}
Again, raters are given options to choose for negative answers.

This offline evaluation strategy offers scalability and minimal rater training. However, an online, interactive evaluation infrastructure would allow us to evaluate the ability of the model to handle errors in its own output (from previous predictions) and its robustness while dealing with novel inputs. Future evaluation will be carried out on this new infrastructure.

\subsection{Results}

Comparing the “makes-sense” scores for model-generated vs. human-generated responses, a clear pattern of improvement emerges based on dataset size. As shown in Table \ref{table:size}, when 5K and 7.5K dialogs are used for the training set, scores for model-generated responses lag behind the human-generated scores by up to 5.5\%. At 10K dialogs, the response scores differ by less than 2\% and model-generated API predictions outperform human labels by 2.5\%. At 21K dialogs, model-generated responses improve to near human-level performance. The 21K model’s API call prediction fares better than human API labeling. As an automatic metric, we also provide the BLEU score generated for each model.

\begin{table}[t]
\setlength{\tabcolsep}{4pt}
\small
\renewcommand{\arraystretch}{1.3}
  \centering
  \begin{tabular}{llll}
\hline \textbf{Size} & \textbf{Plain Resp.} & \textbf{Resp. to API} & \textbf{API call}\\ \hline
\textbf{5K} & & &\\
model: & 86.9\% \textbf{-5.5\%} & 92.3\% \textbf{-3.9\%} & 95.2\% \textbf{-2.2\%}\\
human: & 92.4\% & 96.2\% & 97.4\% \\
BLEU: & 56 \\
\\
\textbf{7.5K} & & &\\
model: & 87.8\% \textbf{-3\%} & 93.8\% \textbf{-2.4\%} & 95.2\% \textbf{-2.3\%}\\
human: & 90.8\% & 96.2\% & 97.7\% \\
BLEU: & 59 \\
\\
\textbf{10K} & & &\\
model: & 86.5\% \textbf{-1.9\%} & 91.8\% \textbf{-1.4\%} & 97.1\% \textbf{+2.5\%}\\
human: & 88.4\% & 93.2\% & 94.6\% \\
BLEU: & 61 \\
\\
\textbf{21K} & & &\\
model: & 89.8\% \textbf{-1.4\%} & 95.3\% \textbf{-0.3\%} & 93.9\% \textbf{+0.3\%}\\
human: & 91.2\% & 95.6\% & 93.6\% \\
BLEU: & 60 \\
\hline\hline
\multicolumn{4}{c}{\textbf{No Pre-training}} \\

\textbf{10K} & & &\\
model: & 55.8\% \textbf{-32.6\%} & 63.1\% \textbf{-30.1\%} & 72.8\% \textbf{-21.8\%}\\
BLEU: & 51 \\
\hline
  \end{tabular}
  \caption{\small{Effects of training set size and pre-training on model accuracy}}
  \label{table:size}
\end{table}

The effect of pre-training is also very clear. After training a fifth model, this time without the T5 framework's pre-training feature, we see a huge drop in evaluation scores. As shown at the bottom of Table \ref{table:size}, we see a decrease of 30\% in model performance for verbal responses and about a 25\% drop in API call prediction accuracy. 

\begin{figure}[h!]
\centering
\includegraphics[width=0.48\textwidth]{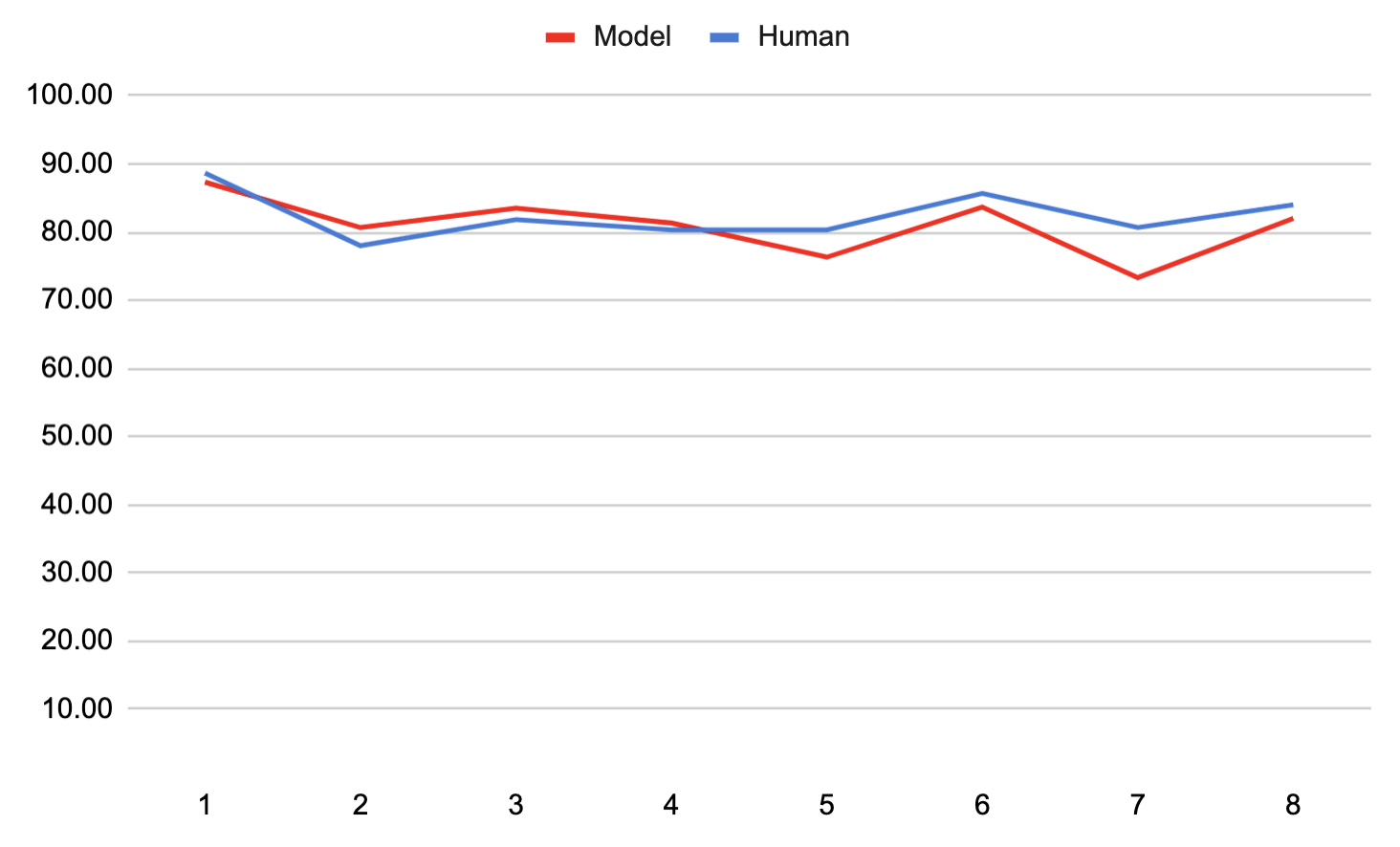}
\caption{\small{Model accuracy per dialog exchange}}
\label{fig:spans-dialog}
\end{figure}

Finally, the quality of the model's prediction stays on par with human scores throughout the conversation as the context grows. Figure \ref{fig:spans-dialog} shows how the model's "makes sense" score stay on the same path after each exchange.

\section{Conclusion}

We have described an end-to-end dialog system approach that shows promising potential for transaction-based dialog applications. In offline human evaluations, our single-domain models trained on just 10,000 dialogs generate responses and predict API calls with near-human level accuracy. One key aspect of this strategy is combining natural language output and structured API calls into a unified text-to-text format in order to leverage general purpose text-to-text transformers, such as the T5 framework. In this way, predicting which API call to make and when is essentially the same as generating the appropriate utterance at a given point in the conversation. The pre-training component significantly boosts performance on our downstream task of fine tuning models on the our datasets. These carefully curated and sufficiently large datasets are also core to this strategy, and creating them is straightforward using the self-dialog technique and simple, API-focused annotation. The TicketTalk dataset released with this paper is one such example. When compared with more traditional, modular system architectures, our end-to-end approach should significantly reduce design and engineering time and resources needed to build task-based dialog systems. Future work will include interactive evaluation of current models as well as an application of this approach to multiple-domain systems.

\section*{Acknowledgments}

We would like to thank our colleagues Daniel De Freitas Adiwardana, Noam Shazeer, Filip Radlinksi, and Pedro Moreno for their discussion and insights through several iterations of this paper. We thank Hadar Shemtov for his guidance and support of the overall project.

\bibliography{main}
\bibliographystyle{acl_natbib}

\end{document}